\documentclass[10pt,twocolumn,letterpaper]{article}

\usepackage{cvpr}
\usepackage{times}
\usepackage{epsfig}
\usepackage{graphicx}
\usepackage{subfigure}
\usepackage{amsmath}
\usepackage{amssymb}
\usepackage{float}
\usepackage{tabularx}

\newcommand{\myparagraph}[1]{\vspace{0.1em}\noindent\textbf{#1}}

\cvprfinalcopy 

\usepackage[pagebackref=true,breaklinks=true,letterpaper=true,colorlinks,bookmarks=false]{hyperref}

\usepackage[font=footnotesize, labelfont=footnotesize, labelfont=bf]{caption}

\def\cvprPaperID{****} 

\begin{document}




\title{EventCap: Monocular 3D Capture of High-Speed Human Motions \\ using an Event Camera}

\author{Lan Xu\textsuperscript{1,2,3} \;\, Weipeng Xu\textsuperscript{2} \;\, Vladislav Golyanik\textsuperscript{2} \;\, Marc Habermann\textsuperscript{2} \;\, Lu Fang\textsuperscript{1} \;\, Christian Theobalt\textsuperscript{2}}

\makeatletter
\let\@oldmaketitle\@maketitle
\renewcommand{\@maketitle}{
	\@oldmaketitle
	\centering
	\vspace{-8mm}
	{\large \textsuperscript{1}Tsinghua-Berkeley Shenzhen Institute, Tsinghua University, China}\\
	{\large	\textsuperscript{2}Max Planck Institute for Informatics, Saarland Informatics Campus, Germany}\\
	{\large	\textsuperscript{3}Robotics Institute, Hong Kong University of Science and Technology, Hong Kong}
	\vspace{8mm}
}
\makeatother

\maketitle

\begin{abstract} 
   The high frame rate is a critical requirement for capturing fast human motions. 
   In this setting, existing markerless image-based methods are constrained by the lighting requirement, the high data bandwidth and the consequent high computation overhead. 
   In this paper, we propose EventCap --- the first approach for 3D capturing of high-speed human motions using a single event camera. 
   Our method combines model-based optimization and CNN-based human pose detection to capture high-frequency motion details and to reduce the drifting in the tracking.
   As a result, we can capture fast motions at millisecond resolution with significantly higher data efficiency than using high frame rate videos. 
   Experiments on our new event-based fast human motion dataset demonstrate the effectiveness and accuracy of our method, as well as its robustness to challenging lighting conditions.
\end{abstract}

\section{Introduction}

\begin{figure}[tbp] 
	\centering 
	\includegraphics[width=1\linewidth]{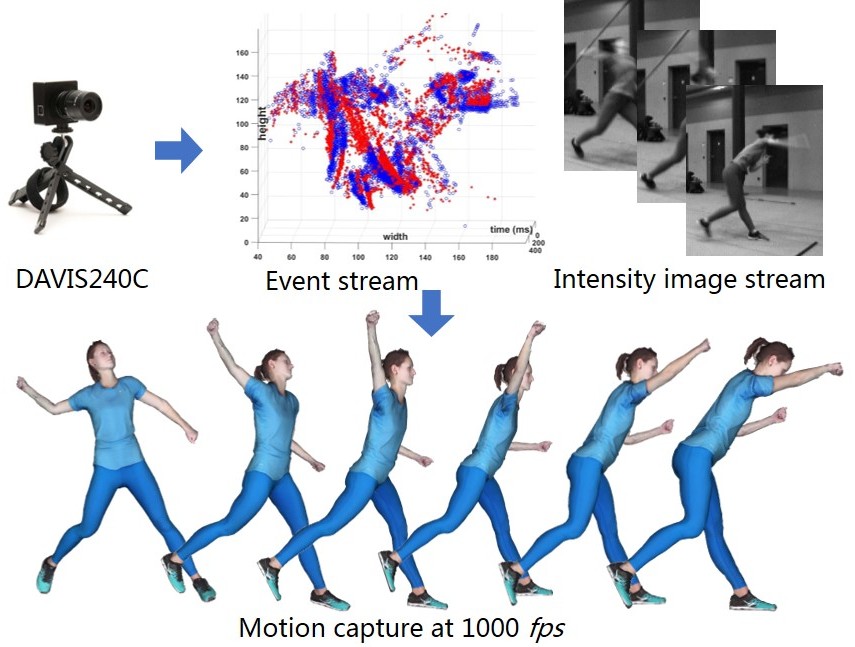} 
	\caption{We present the first monocular event-based 3D human motion capture approach. Given the event stream and the low frame rate intensity image stream from a single event camera, our goal is to track the high-speed human motion at 1000 frames per second.
	} 
	\label{fig:fig_1_teaser} 
	\vspace{-8pt} 
\end{figure} 

With the recent popularity of virtual and augmented reality (VR and AR), there has been a growing demand for reliable 3D human motion capture. 
As a low-cost alternative to the widely used marker and sensor-based solutions, markerless video-based motion capture alleviates the need for intrusive body-worn motion sensors and markers. 
This research direction has received increased attention over the last years~\cite{Davison2001, hasler2009markerless, StollHGST2011, wang2017outdoor, MonoPerfCap}. 

In this paper, we focus on markerless motion capture for high-speed movements, which is essential for many applications such as training and performance evaluation for gymnastics, sports and dancing.
Capturing motion at a high frame rate leads to a very high data bandwidth and algorithm complexity for the existing methods.
While the current marker and sensor-based solutions can support more than $400$ frames per second (\textit{fps})~\cite{VICON, XSENS, phasespace}, the literature on markerless high frame rate motion capture is sparse.

Several recent works \cite{Kowdle2018the, yuan2019temporal} revealed the importance of the high frame rate camera systems for tracking fast motions. 
However, they still suffer from the aforementioned fundamental problem --- the high frame rate leads to excessive amounts of raw data and large bandwidth requirement for data processing (\textit{e.g.,} capturing RGB stream of VGA resolution at $1000$ \textit{fps} from a single view for one minute yields $51.5$GB of data). 
Moreover, both methods~\cite{Kowdle2018the,yuan2019temporal} assume 1) well-lit scenarios for compensating the short exposure time at high frame rate, and 2) indoor capture due to the limitation of the IR-based depth sensor.

In this paper, we propose a rescue to the problems outlined above by using an event camera.
Such bio-inspired dynamic vision sensors~\cite{Lichtsteiner2008} asynchronously measure per-pixel intensity changes and have multiple advantages over conventional cameras, including high temporal resolution, high dynamic range ($140$dB), low power consumption and low data bandwidth. 
These properties potentially allow capturing very fast motions with high data efficiency and in general lighting conditions.
Nevertheless, using the event camera for motion capture is still challenging. 
First, the high temporal resolution of the algorithm leads to very sparse measurements (events) in each frame interval, since the inter-frame intensity changes are subtle. 
The resulting low signal-to-noise ratio (SNR) makes it difficult to track the motion robustly. 
Second, since the event stream only encodes temporal intensity changes, it is difficult to initialize the tracking and prevent drifting. 
A na\"{i}ve solution is to reconstruct images at a high frame rate by accumulating the events and apply existing methods on the reconstructed images. 
Such a policy makes the data dense again, and the temporal information encoded in the events is lost. 

To tackle these challenges, we propose \textit{EventCap} -- the first monocular event-based 3D human motion capture approach (see Fig.~\ref{fig:fig_1_teaser} for an overview). 
More specifically, we design a hybrid and asynchronous motion capture algorithm that leverages the event stream and the low frame rate intensity image stream from the event camera in a joint optimization framework.
Our method consists of three stages:
First, we track the events in 2D space in an asynchronous manner and reconstruct the continuous spatio-temporal event trajectories between each adjacent intensity images.
By evenly slicing the continuous event trajectories, we achieve 2D event tracking at the desired high frame rate.
Second, we estimate the 3D motion of the human actor using a batch-based optimization algorithm. 
To tackle drifting due to the accumulation of tracking errors and depth ambiguities inherent to the monocular setting, our batch-based optimization leverages not only the tracked event trajectories but also the CNN-based 2D and 3D pose estimation from the intensity images. 
Finally, we refine the captured high-speed motion based on the boundary information obtained from the asynchronous event stream. 
To summarise, the \textbf{main contributions} of this paper include: 
\begin{itemize} 
	\setlength\itemsep{0em}
	\item We propose the first monocular approach for event camera-based 3D human motion capture. 
	\item To tackle the challenges of low signal-to-noise ratio (SNR), drifting and the difficulty in initialization, we propose a novel hybrid asynchronous batch-based optimization algorithm. 
	\item We propose an evaluation dataset for event camera-based fast human motion capture and provide high-quality motion capture results at $1000$ \textit{fps}. The dataset will be publicly available.
\end{itemize} 

\begin{figure*}[t] 
	\begin{center} 
		\includegraphics[width=1.0\linewidth]{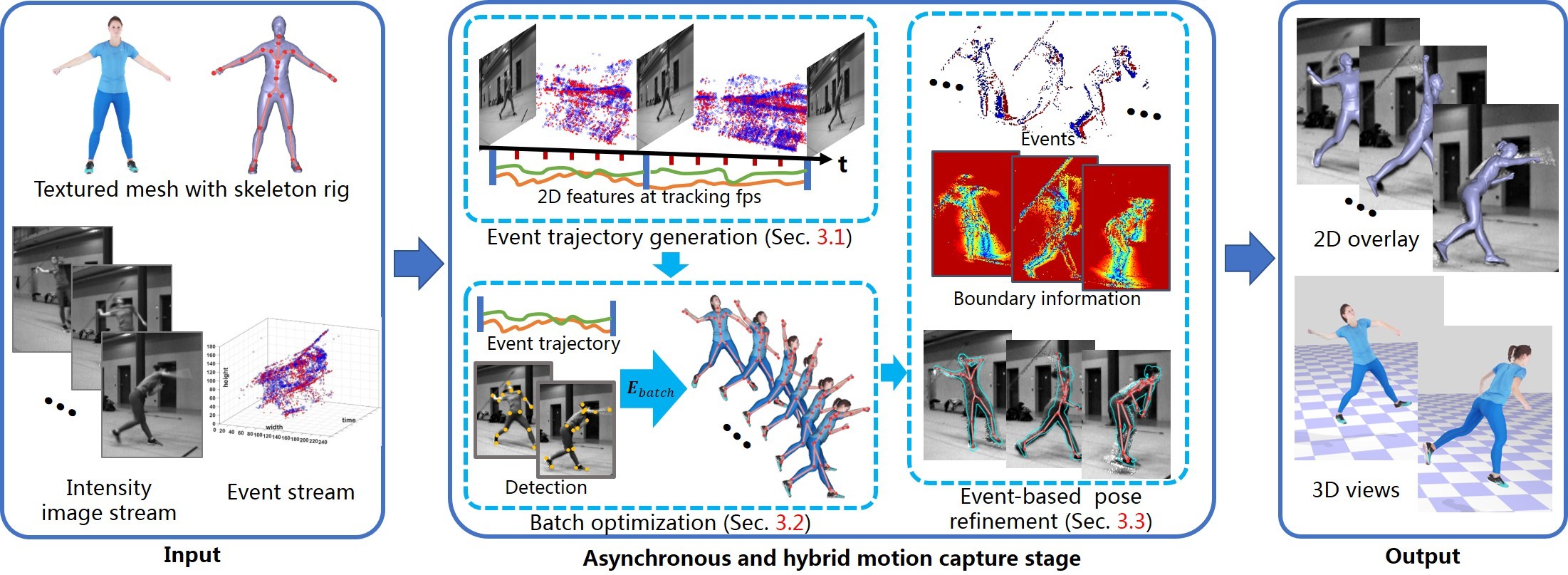} 
	\end{center} 
	\vspace{-4ex}
	\caption{The pipeline of EventCap for accurate 3D human motion capture at a high frame rate. Assuming the hybrid input from a single event camera and a personalized actor rig, we first generate asynchronous event trajectories (Sec.~\ref{sec:trajectory}). Then, the temporally coherent per-batch motion is recovered based on both the event trajectories and human pose detections (Sec.~\ref{sec:batch}). Finally, we perform event-based pose refinement (Sec.~\ref{sec:silouette}).} 
	\label{fig:fig_2_overview} 
\end{figure*} 

\vspace{0.25cm}

\section{Related Work} 

\myparagraph{3D Human Motion Capture.} 
Marker-based multi-view motion capture studios are widely used in both industry and academia~\cite{XSENS, VICON, phasespace}, which can capture fast motions at high frame rate (\textit{e.g.,} $960$ \textit{fps})~\cite{phasespace}. 
Those systems are usually costly, and it is quite intrusive for the users to wear the marker suites. 
Markerless multi-camera motion capture algorithms overcome these problems  \cite{BreglM1998,TheobASST2010,MoeslHKS2011,HolteTTM2012,Gall:2010,SigalBB2010,SigalIHB2012,StollHGST2011,JooLTGNMKNS2015,UnstructureLan}. 
Recent work \cite{AminARS2009,BurenSC2013,ElhayAJTPABST2015,RhodiRRST2015,Robertini:2016,Pavlakos17,Simon17} even demonstrates robust out-of-studio motion capture. 
Although the cost is drastically reduced, synchronizing and calibrating multi-camera systems is still cumbersome. 
Furthermore, when capturing fast motion at high frame rate~\cite{Kowdle2018the}, a large amount of data from multiple cameras becomes a bottleneck  not only for the computation  but also for data processing and storage. 

The availability of commodity depth cameras enabled low-cost motion capture without complicated multi-view setups  \cite{Shotton:2011,Baak:2011,Wei:2012,DoubleFusion,guoTwinFusion}. 
To capture fast motions, Yuan~\textit{et al.}~\cite{yuan2019temporal} combine a high frame rate action camera with a commodity $30$ \textit{fps} RGB-D camera, resulting in a synthetic depth camera of $240$ \textit{fps}. 
However, the active IR-based cameras are unsuitable for outdoor capture, and their high power consumption limits the mobile application.

Recently, purely RGB-based monocular 3D human pose estimation methods have been proposed with the advent of deep neural networks \cite{Ionescu14a, Rogez16, ChenWLSTLCC2016, varol17,  Kovalenko2019arXiv}. 
These methods either regress the root-relative 3D positions of body joints from single images \cite{Li14a,tekin_structured_bmvc16,Zhou16b,mono-3dhp2017,Tekin17a,pavlakos17volumetric,Mehta2017}, or lift 2D detection to 3D \cite{Bogo16,Zhou16a,Chen2016,Yasin16,Jahangiri17}.
The 3D positional representation used in those works is not suitable for animating 3D virtual characters. 
To solve this problem, recent works regress joint angles directly from the images \cite{hmrKanazawa17, Kolotouros_2019_CVPR,NBF:3DV:2018,pavlakos2018humanshape,tan2018indirect}.
In theory, these methods can be applied directly on high frame rate video for fast motion capture. 
In practice, the tracking error is typically larger than the inter-frame movements, which leads to the loss of fine-level motion details. 
Methods combining data-driven 3D pose estimation and image-guided registration alleviate this problem and can achieve higher accuracy \cite{MonoPerfCap,Habermann:2019:LRH:3313807.3311970}.
However, data redundancy is still an issue. 

Furthermore, when capturing a high frame rate RGB video, the scene has to be well-lit, since the exposure time cannot be longer than the frame interval.
Following \cite{MonoPerfCap}, we combine data-driven method with batch optimization. 
Differently, instead of using high frame rate RGB video, we leverage the event stream and the low frame rate intensity image stream from an event camera.
Compared to RGB-based methods, our approach is more data-efficient and works well in a broader range of lighting conditions. 

\myparagraph{Tracking with Event Cameras.} 
Event cameras are causing a paradigm shift in computer vision, due to their high dynamic range, absence of motion blur and low power consumption. 
For a detailed survey of the event-based vision applications, we refer to \cite{Gallego2019arxiv}. 
The most closely related settings to ours are found in works on object tracking from an event stream. 

The specific characteristics of the event camera make it very suitable for tracking fast moving objects.
Most of the related works focus on tracking 2D objects like known 2D templates~\cite{Ni2015,Mishra2016}, corners~\cite{Vasco2017} and lines~\cite{EverdingConradt2018}.
Piatkowska \textit{et al.}~\cite{Piatkowska2012} propose a technique for multi-person bounding box tracking from a stereo event camera.
Valeiras \textit{et al.}~\cite{ReverterValeiras2015} track complex objects like human faces with a set of Gaussian trackers connected with simulated springs. 

The first 3D tracking method was proposed in \cite{ReverterValeiras2016}, which estimates the 3D pose estimation of rigid objects. 
Starting from a known object shape in a known pose, their method incrementally updates the pose by relating events to the closest visible object edges. 
Recently, Calabrese \textit{et al.}~\cite{DHP19} provide the first event-based 3D human motion capture method based on multiple event cameras.
A neural network is trained to detect 2D human body joints using the event stream from each view.
Then, the 3D body pose is estimated through triangulation.
In their method, the events are accumulated over time, forming image frames as input to the network.
Therefore, the asynchronous and high temporal resolution natures of the event camera are undermined, which prevents the method from being used for high frame rate motion capture.

\section{EventCap Method}\label{sec:algorithm} 

Our goal in this paper is to capture high-speed human motion in 3D using a single event camera. 
In order to faithfully capture the fine-level details in the fast motion, a high temporal resolution is necessary.
Here, we aim at a tracking frame rate of 1000 \textit{fps}.

Fig.~\ref{fig:fig_2_overview} provides an overview of EventCap. 
Our method relies on a pre-processing step to reconstruct a template mesh of the actor.
During tracking, we optimize the skeleton parameters of the template to match the observation of a single event camera, including the event stream and the low frame rate intensity image stream.
Our tracking algorithm consists of three stages:
First, we generate sparse event trajectories between two adjacent intensity images, which extract the asynchronous spatio-temporal information from the event stream (Sec.~\ref{sec:trajectory}). 
Then, a batch optimization scheme is performed to optimize the skeletal motion at 1000 \textit{fps} using the event trajectories and the CNN-based body joint detection from the intensity image stream (Sec.~\ref{sec:batch}). 
Finally, we refine the captured skeletal motion based on the boundary information obtained from the asynchronous event stream (Sec.~\ref{sec:silouette}). 

\myparagraph{Template Mesh Acquisition.}
We use a 3D body scanners~\cite{treedys} to generate the template mesh of the actor.
To rig the template mesh with a parametric skeleton, we fit the Skinned Multi-Person Linear Model (SMPL)\cite{SMPL:2015} to the template mesh by optimizing the body shape and pose parameters, and then transfer the SMPL skinning weights to our scanned mesh.
One can also use image-based human shape estimation algorithms, e.g.~\cite{hmrKanazawa17}, to obtain a SMPL mesh as the template mesh, if the 3D scanner is not available.
A comparison of these two methods is provided in Sec.~\ref{sec:abla}.
To resemble the anatomic constraints of body joints, we reduce the degrees of freedom of the SMPL skeleton.
Our skeleton parameter set $\textbf{S}=[\boldsymbol{\theta}, \textbf{R},\textbf{t}]$ includes the joint angles $\boldsymbol{\theta} \in \mathbb{R}^{27}$ of the $N_J$ joints of the skeleton, the global rotation $ \textbf{R}\in\mathbb{R}^3$ and translation $\textbf{t} \in \mathbb{R}^{3}$ of the root.

\myparagraph{Event Camera Model.} 
Event cameras are bio-inspired sensors that measure the changes of logarithmic brightness $\mathcal{L}({u},t)$ independently at each pixel and provide an asynchronous event stream at microsecond resolution. 
An event $e_i=({u}_i, t_i, \rho_i)$ is triggered at pixel ${u}_i$ at time $t_i$ when the logarithmic brightness change reaches a threshold: 
$\mathcal{L}({u}_i, t_i) - \mathcal{L}({u}_i, t_p) = p_i{C}$,
where $t_p$ is the timestamp of the last event occurred at ${u}_i$, $p_i\in\{-1,1\}$ is the event polarity corresponding to the threshold $ \pm{C}$.
Besides the event stream, the camera also produces an intensity image stream at a lower frame rate, which can be expressed as an average of the latent images during the exposure time: 
\begin{equation} \label{eq:e_image}
\mathcal{I}(k)=\frac{1}{T}\int_{t_k-T/2}^{t_k+T/2}\exp(\mathcal{L}(t))dt,
\end{equation} 
where $t_k$ is the central timestamp of the $k$-th intensity image and $T$ is the exposure time. 
Note that $\mathcal{I}(k)$ can suffer from severe motion blur due to high-speed motions.

\subsection{Asynchronous Event Trajectory Generation} \label{sec:trajectory}
A single event does not carry any structural information and therefore tracking based on isolated events is not robust.
To extract the spatio-temporal information from the event stream, in the time interval $[t_{k},t_{k+1}]$ (denoted as the $k$-th batch) between adjacent intensity images $\mathcal{I}(k)$ and $\mathcal{I}(k+1)$, we use~\cite{Gehrig2018} to track the photometric 2D features in an asynchronous manner, resulting in the sparse event trajectories $\{\mathcal{T}(h)\}$. 
Here, $h\in[1,H]$ denotes the temporal 2D pixel locations of all the $H$ photometric features in the current batch,
which are further utilized to obtain correspondences to recover high-frequency motion details.

\noindent{\bf Intensity Image Sharpening.} 
Note that \cite{Gehrig2018} relies on sharp intensity images for gradient calculation. 
However, the intensity images suffer from severe motion blur due to the fast motion.
Thus, we first adopt the event-based double integral (EDI) model \cite{pan2018bringing} to sharpen the images $\mathcal{I}(k)$ and $\mathcal{I}(k+1)$. 
A logarithmic latent image $\mathcal{L}(t)$ can be formulated as $\mathcal{L}(t) = \mathcal{L}(t_k) + \mathcal{E}(t)$, 
where $\mathcal{E}(t)=\int_{t_k}^{t}p_i(s)C\delta(s)ds$ denotes continuous event accumulation. 
By aggregating the latent image $\mathcal{I}(k)$ (see Eq.~\eqref{eq:e_image}) and the logarithmic intensity changes, we obtain the sharpened image: 
\begin{equation} \label{eq:sharp_2} 
\hspace{-7.32pt}\mathcal{L}(t_k) = \log\big(\mathcal{I}(k)\big) -  \log\bigg(\frac{1}{T}\int_{t_k-T/2}^{t_k+T/2}\exp\big(\mathcal{E}(t)\big)dt\bigg).
\end{equation} 
We extract 2D features from the sharpened images $\mathcal{L}(t_k)$ and $\mathcal{L}(t_{k+1})$ instead of the original blurry images. 

\begin{figure}[t!] 
	\begin{center} 
		\includegraphics[width=1\linewidth]{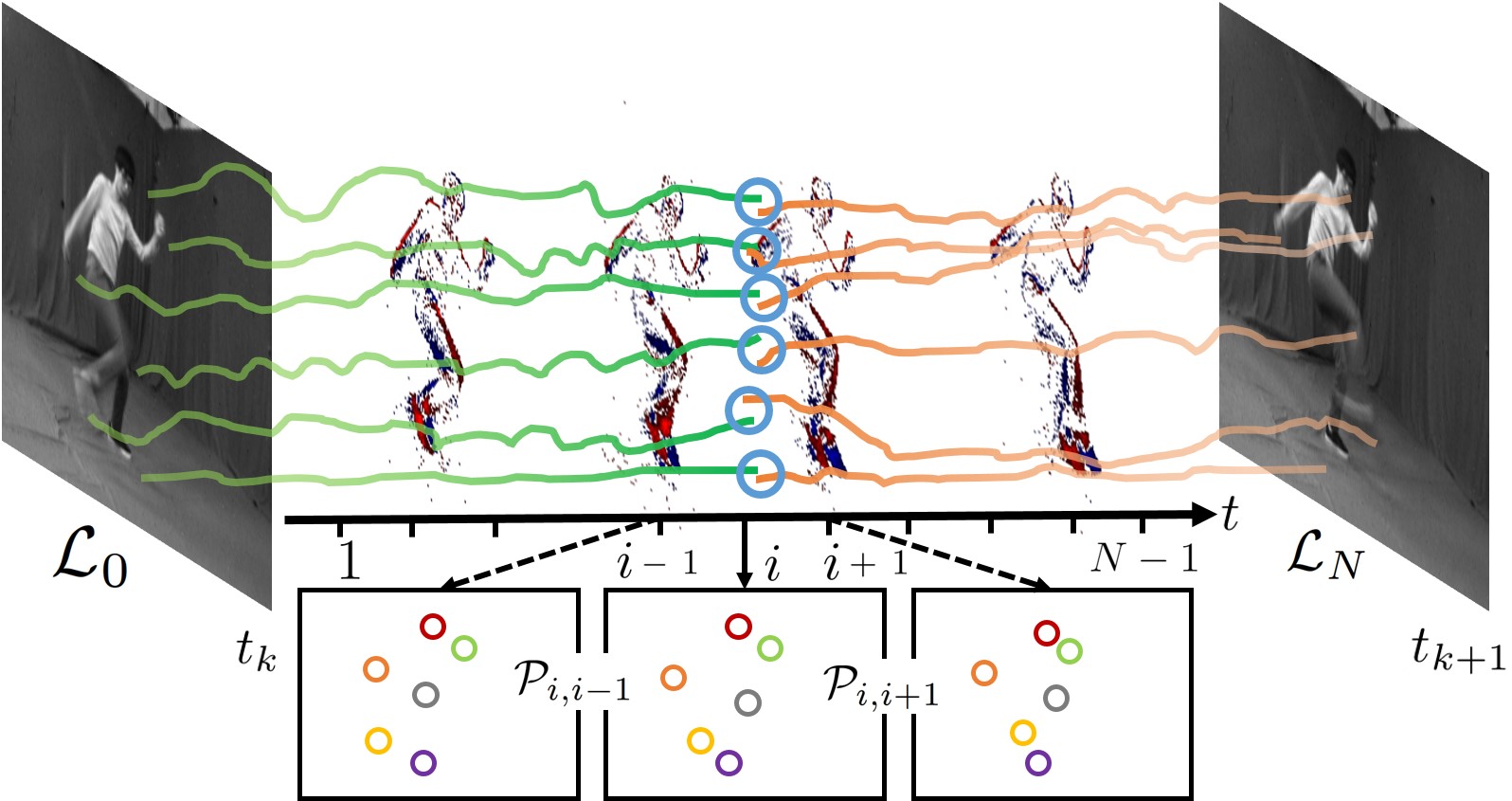} 
	\end{center} 
	\vspace{-15pt}
	\caption{Illustration of asynchronous event trajectories between two adjacent intensity images. The green and orange curves represent the forward and backward event trajectories of exemplary photometric features. The blue circles denote alignment operation. The color-coded circles below indicate the 2D feature pairs between adjacent tracking frames.} 
	\label{fig:fig_3_pairs} 
\end{figure}

\noindent{\bf Forward and Backward Alignment.} The feature tracking can drift over time. To reduce the tracking drifting, we apply the feature tracking method both forward from 
$\mathcal{L}(t_k)$ and backward from $\mathcal{L}(t_{k+1})$. 
As illustrated in Fig.~\ref{fig:fig_3_pairs}, the bidirectional tracking results are stitched by associating the closest backward feature position to each forward feature position at the central timestamp $(t_k+t_{k+1})/2$. 
The stitching is not applied if the 2D distance between the two associated locations is farther than a pre-defined threshold (four pixels). 
For the $h$-th stitched trajectory, we fit a B-spline curve to its discretely tracked 2D pixel locations in a batch and calculate a continuous event feature trajectory $\mathcal{T}(h)$. 

\noindent{\bf Trajectory Slicing.} 
In order to achieve motion capture at the desired tracking frame rate, e.g. 1000 \textit{fps}, we evenly slice the continuous event trajectory $\mathcal{T}(h)$ at each millisecond time stamp (see Fig.~\ref{fig:fig_3_pairs}).
Since we perform tracking on each batch independently, for simplification we omit the subscript $k$ and let $0,1,...,N$ denote the indexes of all the tracking frames for the current batch, where $N$ equals to the desired tracking frame rate divided by the frame rate of the intensity image stream. 
Thus, the intensity images $\mathcal{I}(k)$ and $\mathcal{I}(k+1)$ are denoted as $\mathcal{I}_0$ and $\mathcal{I}_N$ for short, and the corresponding latent images as $\mathcal{L}_0$ and $\mathcal{L}_N$. 

\subsection{Hybrid Pose Batch Optimization} \label{sec:batch} 
Next, we jointly optimize all the skeleton poses $\mathcal{S} = \{\textbf{S}_f\},f\in[0,N]$ for all the tracking frames in a batch.
Our optimization leverages the hybrid input modality from the event camera.
That is, we leverage not only the event feature correspondences obtained in Sec.~\ref{sec:trajectory}, but also the CNN-based 2D and 3D pose estimates to tackle the drifting due to the accumulation of tracking errors and the inherent depth ambiguities of the monocular setting. 
We phrase the pose estimation across a batch as a constrained optimization problem: 
\begin{equation} \label{eq:batch_opt}
\begin{split}
&\mathcal{S}^* = \mathop{\arg\min}_{\mathcal{S}}\boldsymbol{E}_{\mathrm{batch}}(\mathcal{S})\\ 
&\mathrm{s.t.} \quad \boldsymbol{\theta}_{min}\leq  \boldsymbol{\theta}_{f}\leq \boldsymbol{\theta}_{max}, \quad \forall f \in[0,N],
\end{split}
\end{equation} 
where $\boldsymbol{\theta}_{min}$ and $\boldsymbol{\theta}_{max}$ are the pre-defined lower and upper bounds of physically plausible joint angles to prevent unnatural poses. 
Our per-batch objective energy functional consists of four terms: 
\begin{equation} \label{eq:batch}
\begin{split}
\boldsymbol{E}_{\mathrm{batch}}(\mathcal{S}) = &\lambda_{\mathrm{adj}}\boldsymbol{E}_{\mathrm{adj}} + \lambda_{\mathrm{2D}}\boldsymbol{E}_{\mathrm{2D}}+\\
&\lambda_{\mathrm{3D}}\boldsymbol{E}_{\mathrm{3D}} +  \lambda_{\mathrm{temp}}\boldsymbol{E}_{\mathrm{temp}}.
\end{split}
\end{equation} 

\myparagraph{Event Correspondence Term.} 
The event correspondence term exploits the asynchronous spatio-temporal motion information encoded in the event stream. 
To this end, for the $i$-th tracking frame in a batch, we first extract the event correspondences from the sliced trajectories on two adjacent frames $i-1$ and $i+1$, as shown in Fig.~\ref{fig:fig_3_pairs}.
This forms two sets of event correspondences $\mathcal{P}_{i,i-1}$ and $\mathcal{P}_{i,i+1}$, where $\mathcal{P}_{i,*} = \{(p_{i,h},p_{*,h})\},h\in[1,H]$.
The term encourages the 2D projection of the template meshes to match the two sets of correspondences:
\begin{equation} \label{eq:pair}
\small
\hspace{-6.5pt}\boldsymbol{E}_{\mathrm{adj}}(\mathcal{S}) = \sum_{i=1}^{N-1}\sum_{j\in\{i-1, i+1\}}\sum_{h=1}^{H}\tau(p_{i,h})\|\pi(v_{i,h}(\textbf{S}_j))-p_{j,h}\|_2^2,
\end{equation} 
where $\tau(p_{i,h})$ is the indicator which equals to $1$ only if the 2D pixel $p_{i,h}$ corresponds to a valid vertex of the mesh at the $i$-th tracking frame, and $v_{i,h}(\textbf{S}_j)$ is the corresponding vertex on the mesh in pose $\textbf{S}_j$.

\myparagraph{2D and 3D Detection Terms.} 
These terms encourage the posed skeleton to match the 2D and 3D body joint detection obtained by CNN from the intensity images.
To this end, we apply VNect \cite{Mehta2017} and OpenPose~\cite{OpenPose} on the intensity images to estimate the 3D and 2D joint positions, denoted as $\textbf{P}^{3D}_{f,l}$ and $\textbf{P}^{2D}_{f,l}$, respectively, where $f\in\{0,N\}$ is the frame index, and $l$ is the joint index. 
Beside the body joints, We also use the four facial landmarks from the OpenPose~\cite{OpenPose} detection to recover the face orientation. 
The 2D term penalizes the differences between the projection of the landmarks of our model and the 2D detection: 
\begin{equation} \label{eq:2d} 
\boldsymbol{E}_{\mathrm{2D}}(\mathcal{S}) =  \sum_{f\in\{0,N\}}\sum_{l=1}^{N_J+4}\|\pi(J_l(\textbf{S}_f))-\textbf{P}^{2D}_{f,l}\|_2^2, 
\end{equation} 
where $J_l(\cdot)$ returns the 3D position of the $l$-th joint or face marker using the kinematic skeleton, and $\pi\colon\mathbb{R}^{3}\rightarrow\mathbb{R}^{2}$ is the perspective projection operator from 3D space to the 2D image plane. 
Our 3D term aligns the model joints and 3D detection: 
\begin{equation} \label{eq:3d}
\boldsymbol{E}_{\mathrm{3D}}(\mathcal{S}) = \sum_{f\in\{0,N\}}\sum_{l=1}^{N_J}\|J_l(\textbf{S}_f)-(\textbf{P}^{3D}_{f,l}+\textbf{t}')\|_2^2,
\end{equation} 
where $\textbf{t}'\in\mathbb{R}^{3}$ is an auxiliary variable that transforms $\textbf{P}^{3D}_{f,l}$ from the root-centred to the global coordinate system \cite{MonoPerfCap}.

\myparagraph{Temporal Stabilization Term.} 
Since only the moving body parts can trigger events, so far, the non-moving body parts are not constrained by our energy function.
Therefore, we introduce a temporal stabilization constraint for the non-moving body parts.
This term penalizes the changes in joint positions between the current and previous tracking frames:
\begin{equation} \label{eq:temp}
\hspace{-5pt}\boldsymbol{E}_{\mathrm{temp}}(\mathcal{S}) = \sum_{i=0}^{N-1}\sum_{l=1}^{N_{J}}\phi(l)\|J_l(\textbf{S}_i)-J_l(\textbf{S}_{i+1})\|_2^2,
\end{equation} 
where the indicator $\phi(\cdot)$ equals to $1$ if the corresponding body part is not associated with any event correspondence, and equals $0$ otherwise. 

\myparagraph{Optimization.} We solve the constrained optimization problem \eqref{eq:batch_opt} using the Levenberg-Marquardt (LM) algorithm of \textit{ceres} \cite{ceresSolver}. 
For initialization, we minimize the 2D and 3D joint detection terms $\boldsymbol{E}_{\mathrm{2D}}+\boldsymbol{E}_{\mathrm{3D}}$ to obtain the initial values of $\textbf{S}_0$ and $\textbf{S}_N$, and then linearly interpolate $\textbf{S}_0$ and $\textbf{S}_N$ to obtain the initial values of all the tracking frames $\{\textbf{S}_f\}$ in the current batch proportional to their timestamps.

\begin{figure}[t!] 
	\begin{center} 
		\subfigure[]{\includegraphics[height=82pt]{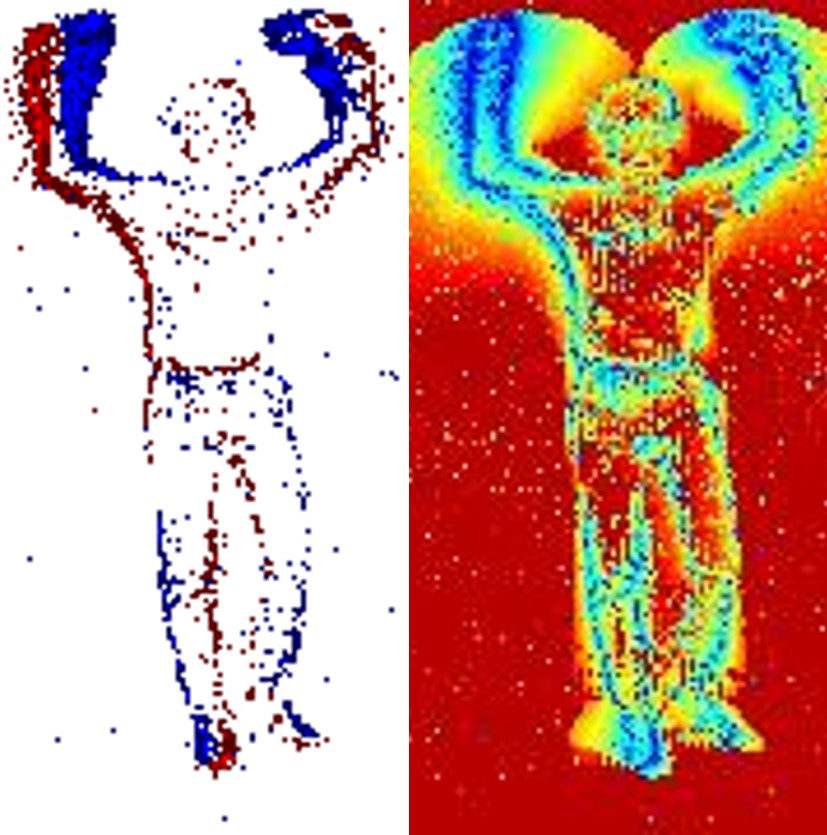}\label{fig:silh_a}}	
		\subfigure[]{\includegraphics[height=82pt]{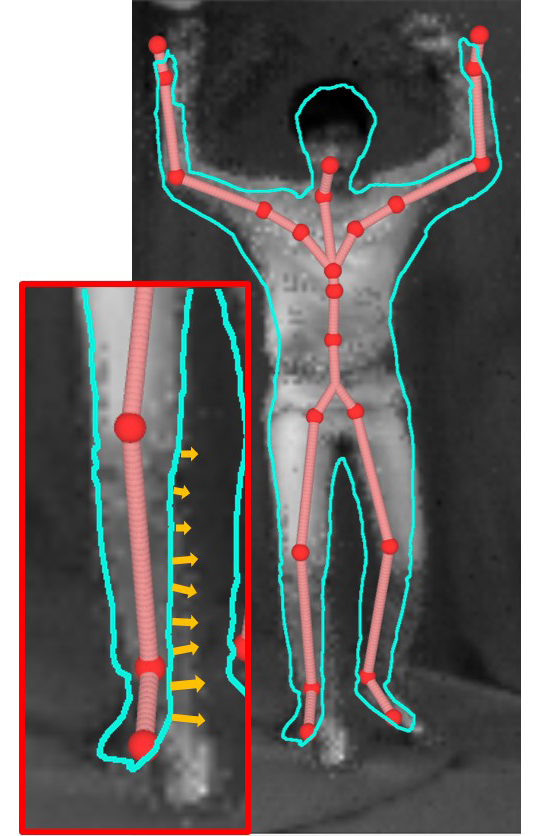}\label{fig:silh_b}}	
		\subfigure[]{\includegraphics[height=82pt]{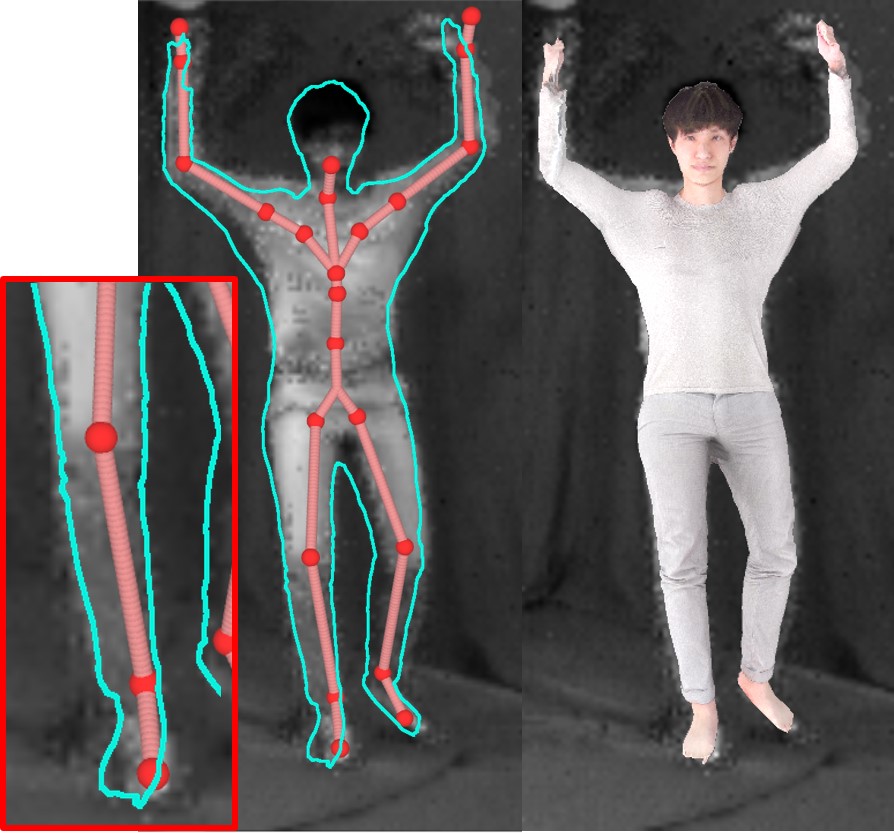}\label{fig:silh_c}}	
	\end{center} 
	\vspace{-5ex}
	\caption{Event-based pose refinement. (a) Polarities and color-coded normalized distance map ranging from 0 (blue) to 1 (red). (b, c) The skeleton overlapped with the latent image before and after the refinement. Yellow arrows indicate the refined boundaries and exemplary 2D correspondences.} 
	\label{fig:fig_3_silhou} 
\end{figure} 

\begin{figure*}[tbp] 
	\begin{center} 
		\includegraphics[width=1\linewidth]{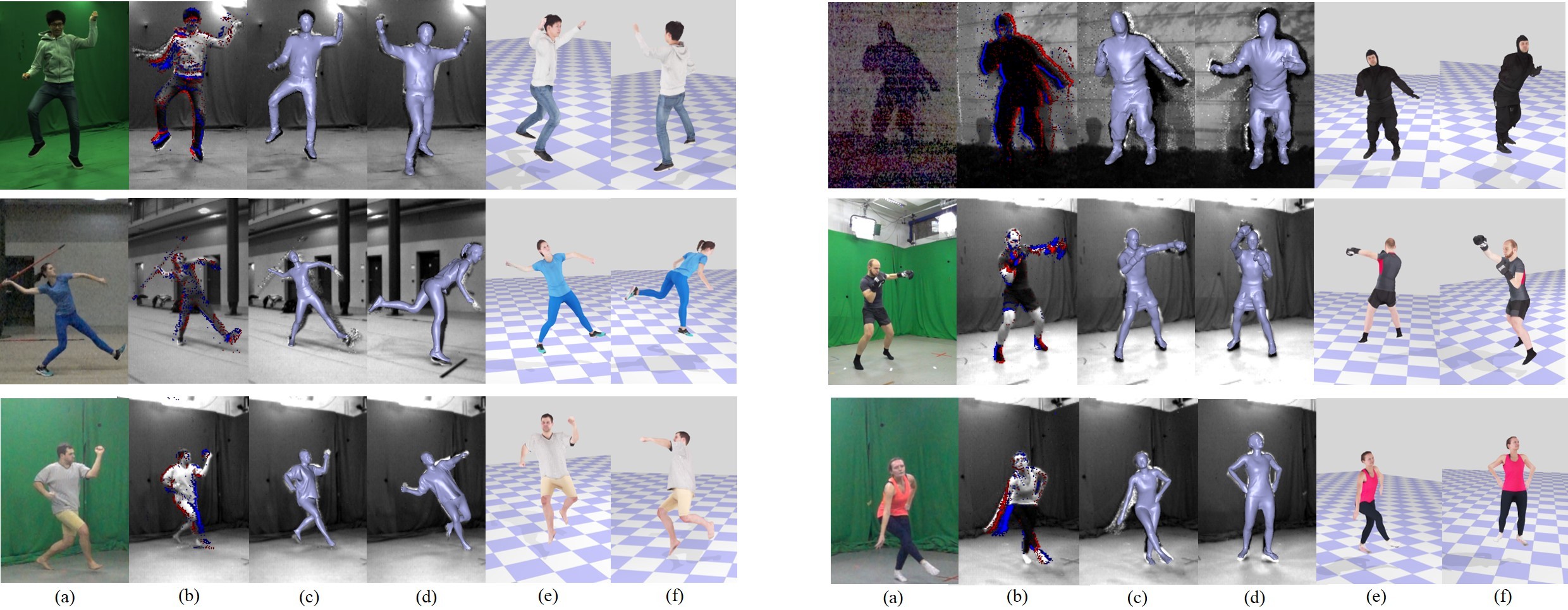} 
	\end{center} 
	\vspace{-18pt}
	\caption{Qualitative results of EventCap on some sequences from our benchmark dataset, including ``wave'', ``ninja'', ``javelin'', ``boxing'', ``karate'' and ``dancing'' from the upper left to lower right. (a) The reference RGB image (not used for tracking); (b) Intensity images and the accumulated events; (c,d) Motion capture results overlaid on the reconstructed latent images; (e,f) Results rendered in 3D views. } 
	\label{fig:fig_all}
\end{figure*}

\subsection{Event-Based Pose Refinement} \label{sec:silouette} 
Most of the events are triggered by the moving edges in the image plane, which have a strong correlation with the actor's silhouette.
Based on this finding, we refine our skeleton pose estimation in an Iterative Closest Point (ICP) \cite{Chen1992} manner.
In each ICP iteration, we first search for the closest event for each boundary pixel of the projected mesh.
Then, we refine the pose $\textbf{S}_f$ by solving the non-linear least squares optimization problem: 
\begin{equation} \label{eq:refine}
\boldsymbol{E}_{\mathrm{refine}}(\textbf{S}_f) = \lambda_{\mathrm{sil}}\boldsymbol{E}_{\mathrm{sil}}(\textbf{S}_f)+ \lambda_{\mathrm{stab}}\boldsymbol{E}_{\mathrm{stab}}(\textbf{S}_f).
\end{equation} 
Here, we enforce the refined pose to stay close to its initial position using the following stability term:
\begin{equation} \label{eq:stable}
\boldsymbol{E}_{\mathrm{stab}}(\textbf{S}_f) = \sum_{l=1}^{N_J}\|J_l(\textbf{S}_f)-J_i(\hat{\textbf{S}}_f)\|_2^2,
\end{equation} 
where $\hat{\textbf{S}}_f$ is the skeleton pose after batch optimization (Sec. \ref{sec:batch}).
The data term $\boldsymbol{E}_{\mathrm{sil}}$ relies on the closest event search, which we will describe later. 
Let $s_b$ and $v_b$ denote the $b$-th boundary pixel and its corresponding 3D position on the mesh based on barycentric coordinates.
For each $s_b$, let $u_b$ denote the corresponding target 2D position of the closest event.
Then $\boldsymbol{E}_{\mathrm{sil}}$ measures the 2D point-to-plane misalignment of the correspondences: 
\begin{equation} \label{eq:silhouette}
\begin{split}
\boldsymbol{E}_{\mathrm{sil}}(\textbf{S}_f) = \sum_{b\in \mathcal{B}}\|\mathbf{n}_b^{\textbf{T}}\big(\pi(v_b(\textbf{S}_f)-u_b)\big)\|_2^2,
\end{split}
\end{equation} 
where $\mathcal{B}$ is the boundary set of the projected mesh and $\mathbf{n}_b\in\mathbb{R}^{2}$ is the 2D normal vector corresponding to $s_b$.

\myparagraph{Closest Event Search.} 
Now we describe how to obtain the closest event for each boundary pixel $s_b$.
The criterion for the closest event searching is based on the temporal and spatial distance between $s_b$ and each recent event $e=(u, t, \rho)$:
\begin{equation} \label{eq:eSil_dist} 
\mathcal{D}(s_b,e) =\lambda_{dist}\|\frac{t_f-t}{t_N-t_0}\|^2_2 + \|s_b-u\|^2_2, 
\end{equation}
where $t_f$ is the timestamp of the current tracking frame, $\lambda_{dist}$ balances the weights of temporal and spatial distances, and $t_N-t_0$ equals to the time duration of a batch.
We then solve the following local searching problem to obtain the closest event for each boundary pixel $s_b$:
\begin{equation} \label{eq:eSil_search} 
e_b=\mathop{\arg\min}_{e\in\mathcal{P}}\mathcal{D}(s_b, e).
\end{equation} 
Here, $\mathcal{P}$ is the collection of events, which happen within a local $8\times8$ spatial patch centred at $s_b$ and within the batch-duration-sized temporal window centered at $t_f$.
The position $u_b$ of the closest event $e_b$ is further utilized in Eq.~\eqref{eq:silhouette}.

\myparagraph{Optimization.} 
During the event-based refinement, we initialize $\textbf{S}_f$ with the batch-based estimates and typically perform four ICP iterations. 
In each iteration, the energy in Eq.~\eqref{eq:refine} is solved using the LM method provided by \textit{ceres} \cite{ceresSolver}. 
As shown in Figs.~\ref{fig:silh_b} and \ref{fig:silh_c}, our iterative refinement based on the event stream improves the pose estimates. 

\vspace{0.35cm}

\section{Experimental Results} 
In this section, we evaluate our EventCap method on a variety of challenging scenarios. 
We run our experiments on a PC with 3.6 GHz Intel Xeon E5-1620 CPU and $16$GB RAM. 
Our unoptimized CPU code takes 4.5 minutes for a batch (i.e. 40 frames or 40ms), which divides to $30$ seconds for the event trajectory generation, $1.5$ minutes for the batch optimization and $2.5$ minutes for the pose refinement. 
In all experiments, we use the following empirically determined parameters:  $\lambda_{3D} = 1$, $\lambda_{2D} = 200$, $\lambda_{adj} = 50$, $\lambda_{temp} = 80$,  $\lambda_{sil} = 1.0$, $\lambda_{stab} = 5.0$, and $\lambda_{dist} = 4.0$. 

\myparagraph{EventCap Dataset.} 
To evaluate our method, we propose a new benchmark dataset for monocular event-based 3D motion capture, consisting of 12 sequences of 6 actors performing different activities, including karate, dancing, javelin throwing, boxing, and other fast non-linear motions.
All our sequences are captured with a DAVIS240C event camera, which produces an event stream and a low frame rate intensity image stream (between 7 and 25 \textit{fps}) at $240 \times 180$ resolution.
For reference, we also capture the actions with a Sony RX0 camera, which produces a high frame rate (between $250$ and $1000$ \textit{fps}) RGB videos at $1920 \times 1080$ resolution.
In order to perform a quantitative evaluation, one sequence is also tracked with a multi-view markerless motion capture system~\cite{captury} at 100 \textit{fps}. 
We will make our dataset publicly available. 

Fig.~\ref{fig:fig_all} shows several example frames of our EventCap results on the proposed dataset. 
For qualitative evaluation, we reconstruct the latent images at 1000 \textit{fps} from the event stream using the method of~\cite{pan2018bringing}.
We can see in Fig.~\ref{fig:fig_all} that our results can be precisely overlaid on the latent images (c-d), and that our reconstructed poses are plausible in 3D (e-f).
The complete motion capture results are provided in our supplementary video.
From the 1000 \textit{fps} motion capture results, we can see that our method can accurately capture the high-frequency temporal motion details, which cannot be achieved by using standard low \textit{fps} videos.
Benefiting from the high dynamic range of the event camera, our method can handle various lighting conditions, even many extreme cases, such as the actor in black ninja suite captured outdoor in the night (see Fig.~\ref{fig:fig_all} top right).
While it is already difficult for human eyes to spot the actor in the reference images, our method still yields plausible results.

\subsection{Ablation Study} \label{sec:abla} 

\begin{figure}[tbp] 
	\begin{center} 
		\includegraphics[width=1\linewidth]{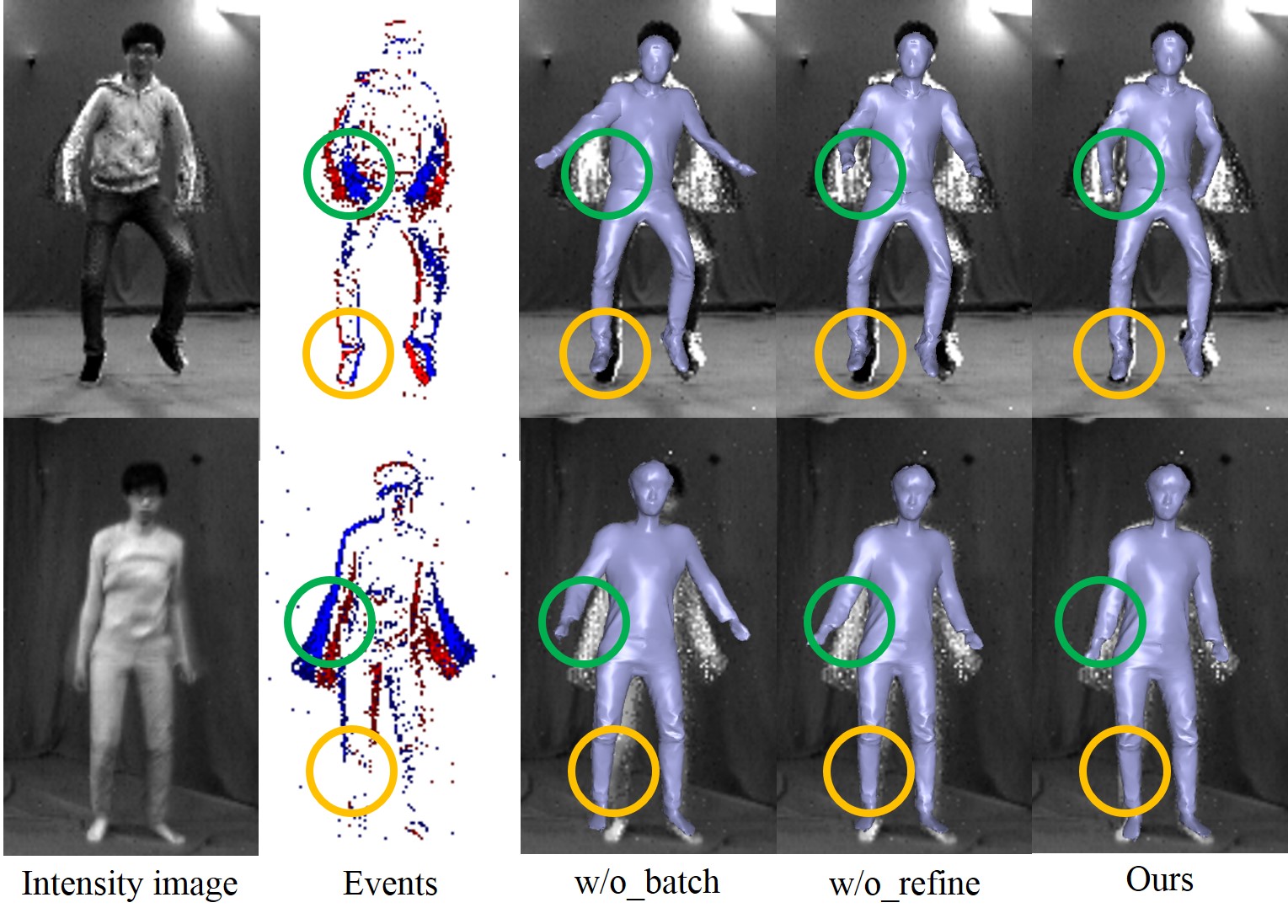} 
	\end{center} 
	\vspace{-18pt}
	\caption{Ablation study for the EventCap components. In the second column, polarity events are accumulated between the time duration from the previous to the current tracking frames. Results of the full pipeline overlay more accurately with the latent images.} 
	\label{fig:fig_comp_abla} 
\end{figure} 
\begin{figure}[tbp] 
	\begin{center} 
		\includegraphics[width=1\linewidth]{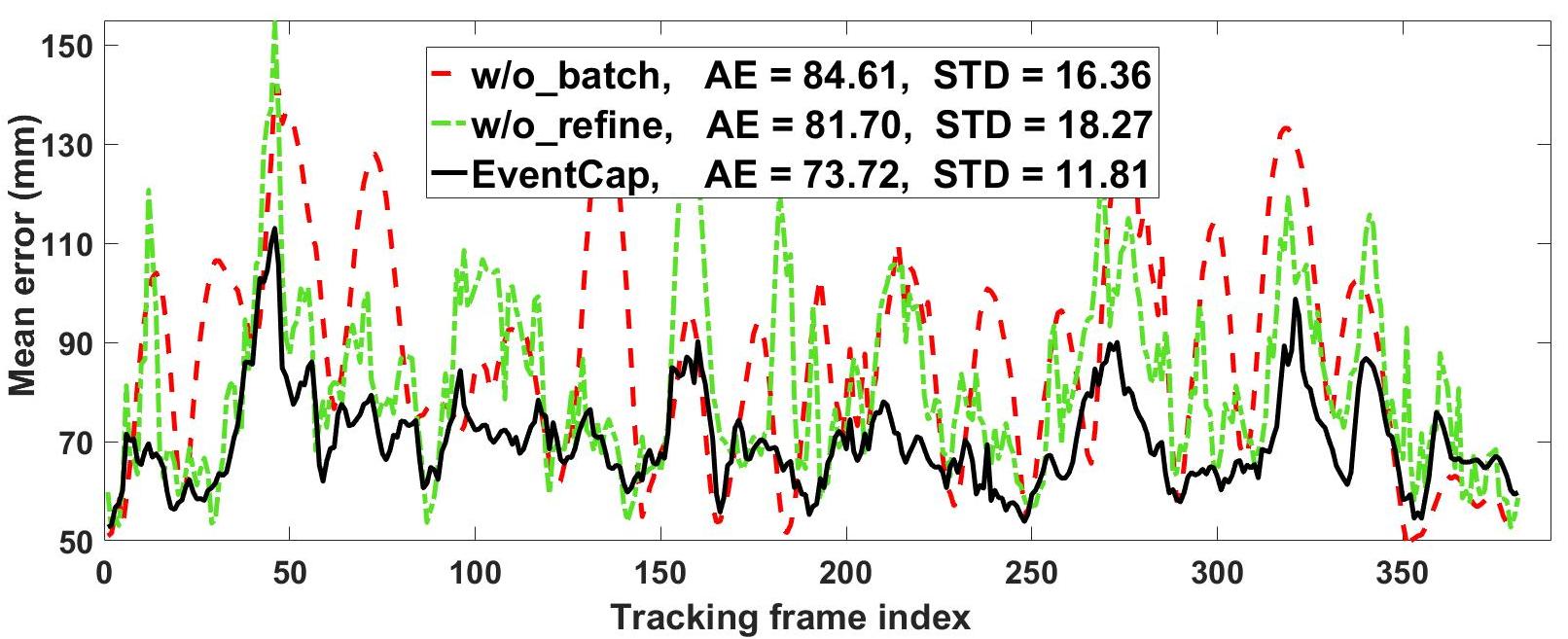} 
	\end{center} 
	\vspace{-12pt} 
	\caption{Ablation study: the average per-joint 3D error demonstrates the effectiveness of each algorithmic component of EventCap. Our full pipeline  consistently achieves the lowest error. }
	\label{fig:fig_abla_curve} 
	\vspace{-8pt} 
\end{figure} 

In this section, we evaluate the individual components of EventCap. 
Let \textit{w/o\_batch} and \textit{w/o\_refine} denote the variations of our method without the batch optimization (Sec.~\ref{sec:batch}) and the pose refinement  (Sec.~\ref{sec:silouette}), respectively. 
For \textit{w/o\_batch}, we optimize the pose for each tracking frame $t\in [0, N]$ independently.
The skeleton poses $\mathcal{S}_t$ are initialized with linear interpolation of the poses obtained from the two adjacent intensity images $\mathcal{I}_0$ and $\mathcal{I}_N$. 
As shown in Fig.~\ref{fig:fig_comp_abla}, the results of our full pipeline are overlaid on the reconstructed latent images more accurately than those of \textit{w/o\_batch} and \textit{w/o\_refine} (the full sequence can be found in our supplementary video).
We can see that --- benefiting from the integration of CNN-based 2D and 3D  pose estimation and the event trajectories --- our batch optimization significantly improves the accuracy and alleviated the drifting problem. 
Our pose refinement further corrects the remaining misalignment, resulting in a better overlay on the reconstructed latent images. 
This is further evidenced by our quantitative evaluation in Fig.~\ref{fig:fig_abla_curve}.

To this end, we obtain ground truth 3D joint positions using a multi-view markerless motion capture method~\cite{captury}.
Then, we compute the average per-joint error (AE) and the standard deviation (STD) of AE on every 10th tracking frame,
because our tracking frame rate is 1000 \textit{fps} while the maximum capture frame rate of~\cite{captury} is 100 \textit{fps}.
Following~\cite{MonoPerfCap}, to factor out the global pose, we perform  Procrustes analysis to rigidly align our results to the ground truth. 
Fig.~\ref{fig:fig_abla_curve} shows our full pipeline consistently outperform the baselines on all frames, yielding both the lowest AE and the lowest STD.
This not only highlights the contribution of each algorithmic component but also illustrates that our approach captures more high-frequency motion details in fast motions and achieves temporally more coherent results.

\begin{figure}[tbp] 
	\begin{center} 
		\includegraphics[width=1\linewidth]{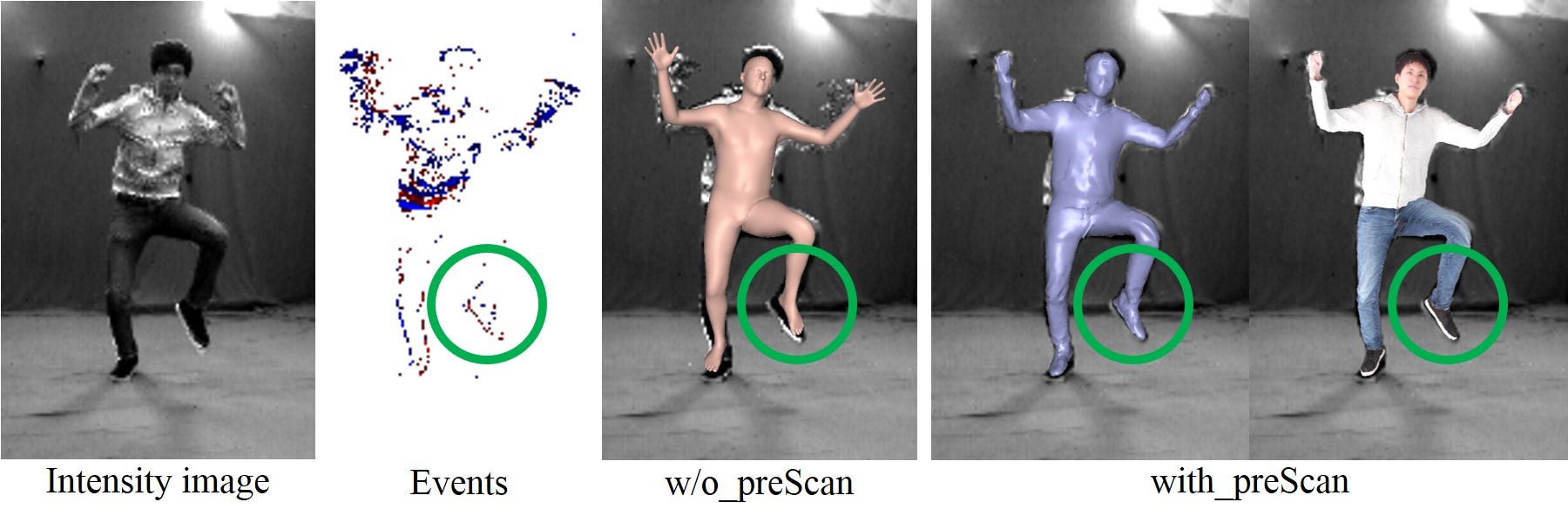} 
	\end{center} 
	\vspace{-18pt}
	\caption{Influence of the template mesh accuracy. Our results using a pre-scanned template and using SMPL mesh are comparable, while the more accurate 3D scanned template improves the overlay on the latent images.} 
	\label{fig:fig_comp_temp} 
\end{figure} 

\begin{figure}[tbp] 
	\begin{center} 
		\includegraphics[width=1\linewidth]{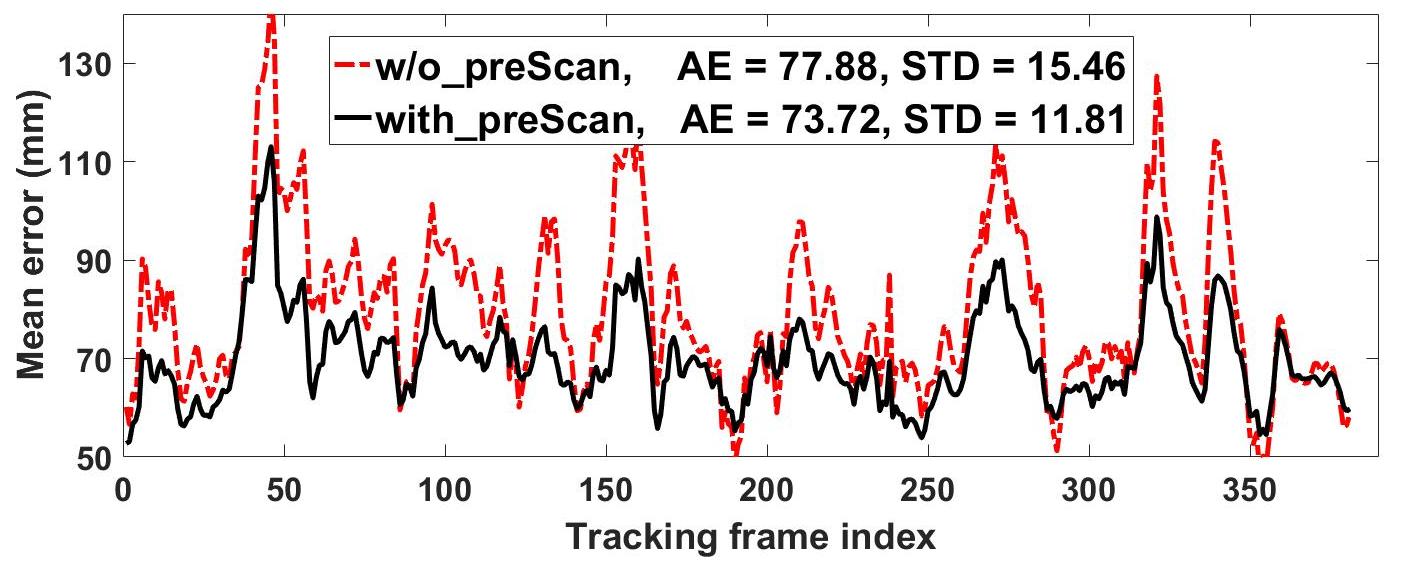} 
	\end{center} 
	\vspace{-18pt} 
	\caption{Quantitative analysis of the template mesh. The more accurate template improves the tracking accuracy in terms of average per-joint error.}
	\label{fig:fig_preScan_curve} 
	\vspace{-8pt} 
\end{figure} 

We further evaluate the influence of the template mesh accuracy. 
To this end, we compare the result using SMPL mesh from image-based body shape estimation~\cite{hmrKanazawa17} (denoted as \textit{w/o\_preScan}) against that using more accurate 3D scanned mesh (denoted as \textit{with\_preScan}).
As shown in Fig.~\ref{fig:fig_comp_temp}, the two methods yield comparable pose estimation results, while the 3D scanned mesh helps in terms of an image overlay since the SMPL mesh cannot model the clothes. 
Quantitatively, the method using 3D scanned mesh achieves a lower AE ($73.72$mm vs $77.88$mm) as shown in Fig.~\ref{fig:fig_preScan_curve}. 
\begin{figure}[tbp] 
	\begin{center} 
		\includegraphics[width=1\linewidth]{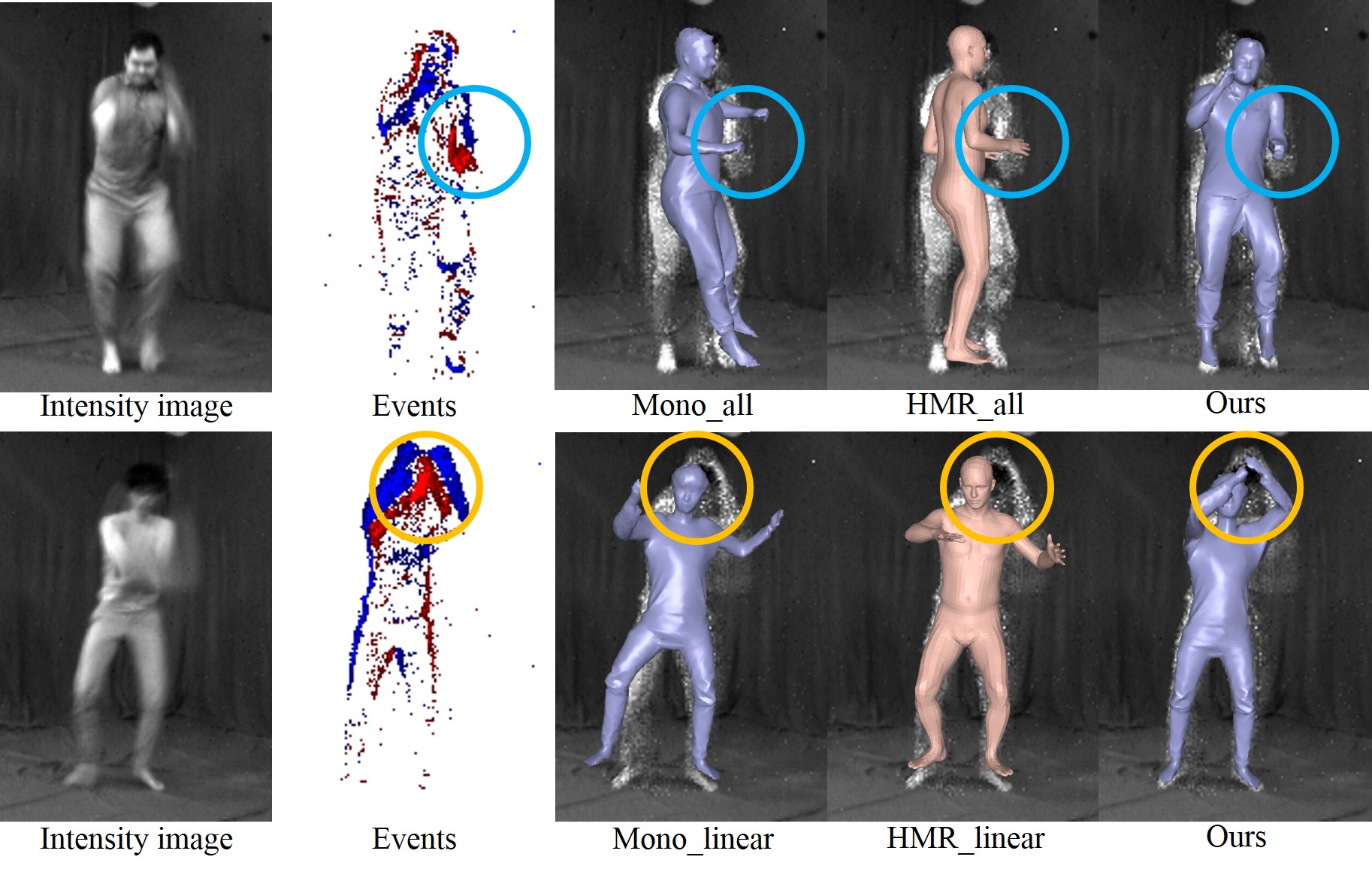} 
	\end{center} 
	\vspace{-4ex}
	\caption{Qualitative comparison. Note that the polarity events are accumulated between the time duration from the previous to the current tracking frames. Our results overlay better with the latent images than the results of other methods.} 
	\label{fig:fig_compare} 
	\vspace{-2ex}
\end{figure}

\subsection{Comparison to Baselines} 
To the best of our knowledge, our approach is the first monocular event-based 3D motion capture method.
Therefore, we compare to existing monocular RGB-based approaches, HMR~\cite{hmrKanazawa17} and MonoPerfCap~\cite{MonoPerfCap}, which are most closely related to our approach.
For a fair comparison, we first reconstruct the latent intensity images at 1000 \textit{fps} using \cite{pan2018bringing}.
Then, we apply HMR \cite{HMR18} and \hbox{MonoPerfCap}\footnote{Only the pose optimization stage of MonoPerfCap is used, as their segmentation does not work well on the reconstructed latent images.}\cite{MonoPerfCap} on all latent images,
denoted as \textit{HMR\_all} and \textit{Mono\_all}, respectively. 
We further apply MonoPerfCap \cite{MonoPerfCap} and HMR \cite{HMR18} only on the raw intensity images of low frame rate and linearly upsample the skeleton poses to 1000 \textit{fps},
denoted as \textit{Mono\_linear} and \textit{HMR\_linear}, respectively. 
As shown in Fig.~\ref{fig:fig_compare}, both \textit{HMR\_all} and \textit{Mono\_all} suffer from inferior tracking results due to the accumulated error of the reconstructed latent images, while \textit{Mono\_linear} and \textit{HMR\_linear} fail to track the high-frequency motions. 
In contrast, our method achieves significantly better tracking results and more accurate overlay with the latent images.
For quantitative comparison, we make use of the sequence with available ground truth poses (see Sec.~\ref{sec:abla}). 
In Table~\ref{table:all}, we report the mean AE of 1) all tracking frames (\textbf{AE\_all}), 2) only the raw intensity frames (\textbf{AE\_raw}), and 3) only the reconstructed latent image frames (\textbf{AE\_nonRaw}).
We also report the data throughput as the size of processed raw data per-second (\textbf{Size\_sec}) for different methods. 
These quantitative results illustrate that our method achieves the highest tracking accuracy in our high frame rate setting.
Furthermore, our method uses only $3.4\%$ of the data bandwidth required in the high frame rate images setting (\textit{HMR\_all} and \textit{Mono\_all}), or only $10\%$ higher compared to the low frame rate upsampling setting (\textit{Mono\_linear} and \textit{HMR\_linear}).

\begin{table}[tpb]
	\footnotesize
	\newcommand{\tabincell}[2]{\begin{tabular}{@{}#1@{}}#2\end{tabular}}
	\newcolumntype{C}[1]{>{\centering\arraybackslash}p{#1}}
	\renewcommand{\arraystretch}{1.2}	
	\centering
	\begin{tabular}{|C{1.2cm}|C{1.3cm}|C{1.3cm}|C{1.3cm}||C{0.7cm}|}
		\hline
		& {\scriptsize{AE\_all (mm)}} &  \scriptsize{AE\_raw (mm)}  & \scriptsize{AE\_nonRaw (mm)}  & \scriptsize{Size\_sec (MB)} \\
		\hline
		\hline
		{\scriptsize{\textit{Mono\_linear}}}& 88.6$\pm${17.3}  & 89.2$\pm${19.7} & 88.5$\pm${16.8} & \textbf{1.83}  \\
		\hline
		{\scriptsize{\textit{Mono\_all}}}& 98.4$\pm$22.8   & 90.2$\pm${21.4} & 99.8$\pm${23.0} & 58.59 \\
		\hline
		{\scriptsize{\textit{HMR\_linear}}}& 105.3$\pm$19.2  & 104.3$\pm${20.6} & 105.4$\pm${19.1} & \textbf{1.83}  \\		
		\hline
		{\scriptsize{\textit{HMR\_all}}}& 110.3$\pm$20.4  & 105.5$\pm${19.5} & {105.4$\pm${20.4}} & 58.59  \\
		\hline
		{\scriptsize{\textbf{Ours}}}& \textbf{73.7$\pm$11.8}  & \textbf{75.2$\pm$13.3} & \textbf{73.5$\pm$11.3} & 2.02  \\			
		\hline
	\end{tabular}
	\vspace{-2ex}
	\caption{ Quantitative comparison of several methods in terms of tracking accuracy and data throughput. }
	\label{table:all} 
\end{table}

\begin{figure}[tbp] 
	\begin{center} 
		\includegraphics[width=1\linewidth]{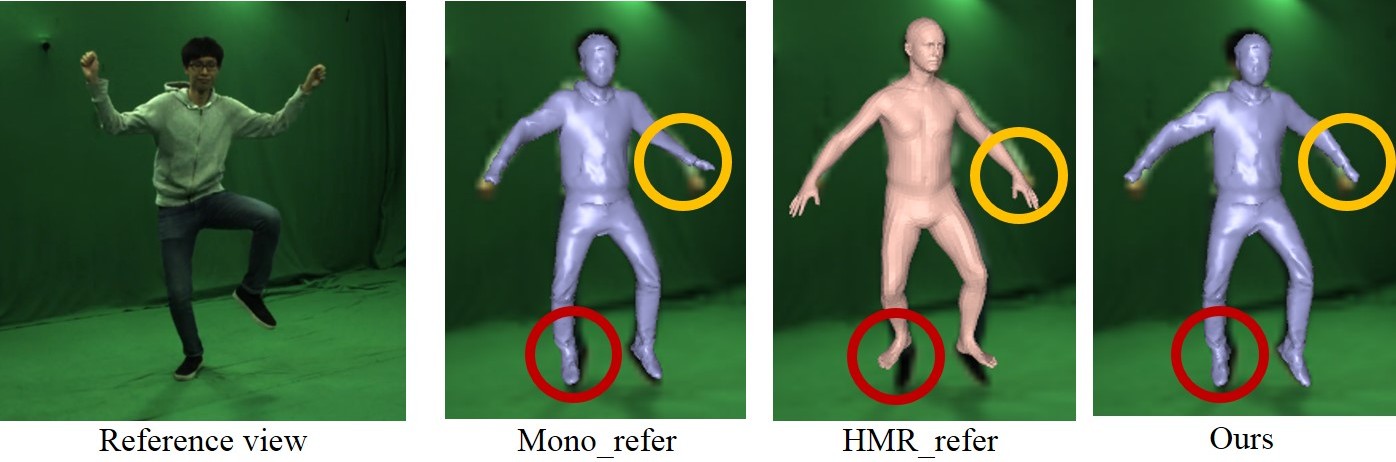} 
	\end{center} 
	\vspace{-4ex}
	\caption{Qualitative comparison. Our results yield similar and even better overlay with the reference image, compared to results of \textit{Mono\_refer} and \textit{HMR\_refer}, respectively.} 
	\label{fig:fig_comp_refer} 
\end{figure}

For further comparison, we apply MonoPerfCap \cite{MonoPerfCap} and HMR \cite{HMR18} to the high frame rate reference images directly, denoted as \textit{HMR\_refer} and \textit{Mono\_refer}, respectively.
Due to the difference of image resolution between the reference and the event cameras, for a fair comparison, we downsample the reference images into the same resolution of the intensity image from the event camera. 
As shown in Fig.~\ref{fig:fig_comp_refer}, our method achieves similar overlap to the reference image without using the high frame rate reference images. 
The corresponding AE and STD for all the tracking frames, as well as the \textit{Size\_sec} are reported in Table \ref{table:large}. 
Note that our method relies upon only $3.4\%$ of the data bandwidth of the reference image-based methods, and even achieves better tracking accuracy compared to \textit{Mono\_refer} and \textit{HMR\_refer}.

\begin{table}[tpb]
	\footnotesize
	\newcommand{\tabincell}[2]{\begin{tabular}{@{}#1@{}}#2\end{tabular}}
	\newcolumntype{C}[1]{>{\centering\arraybackslash}p{#1}}
	\renewcommand{\arraystretch}{1.2}	
	\centering
	\begin{tabular}{|C{1.2cm}|C{1.7cm}|C{1.7cm}|C{1.7cm}|}
		\hline
		& \scriptsize{AE\_all (mm)} &\scriptsize{STD (mm)}& \scriptsize{Size\_sec (MB)} \\
		\hline
		\hline
		{\scriptsize{\textit{Mono\_refer}}}& {76.5} & {13.4} & 58.59 \\
		\hline
		{\scriptsize{\textit{HMR\_refer}}}& 83.5 & 17.8 & 58.59  \\
		\hline
		{\scriptsize{\textbf{Ours}}}& \textbf{73.7} &  \textbf{11.8}  & \textbf{2.02}  \\			
		\hline
	\end{tabular}
	\vspace{-2ex}
	\caption{ Quantitative comparison against \textit{Mono\_refer} and \textit{HMR\_refer} in terms of tracking accuracy and data throughput.}
	\label{table:large} 
	\vspace{-2ex}
\end{table}

\section {Discussion and Conclusion} 
We present the first approach for markerless 3D human motion capture using a single event camera and a new dataset with high-speed human motions. 
Our batch optimization makes full usage of the hybrid image and event streams,
while the captured motion is further refined with a new event-based pose refinement approach. 
Our experimental results demonstrate the effectiveness and robustness of \hbox{EventCap} in capturing fast human motions in various scenarios.
We believe that it is a significant step to enable markerless capturing of high-speed human capture, with many potential applications in 
AR and VR, gaming, entertainment and performance evaluation for gymnastics, sports and dancing. 
In future work, we intend to investigate handling large occlusions and topological changes (\textit{e.g.,}~opening a jacket) and improve the runtime performance.

{\small
\bibliographystyle{ieee}
\bibliography{egbib}
}

\end{document}